\documentclass[10pt,twocolumn,letterpaper]{article}

\usepackage{iccv}
\usepackage{times}
\usepackage{epsfig}
\usepackage{graphicx}
\usepackage{amsmath}
\usepackage{amssymb}
\usepackage{duckuments}
\usepackage{multirow}
\usepackage{tabularx,booktabs}
\usepackage[dvipsnames,table]{xcolor}
\usepackage{pifont}
\usepackage{subfigure}

\usepackage[numbers]{natbib}

\usepackage[breaklinks=true,bookmarks=false]{hyperref}

\iccvfinalcopy 

\def\iccvPaperID{6955} 

\ificcvfinal\fi

\begin{document}

\title{Graph Decision Transformer}

\author{
Shengchao Hu\textsuperscript{\rm 1}
\quad 
Li Shen\textsuperscript{\rm 2}\textsuperscript{\rm}\thanks{Corresponding author: Li Shen}
\quad 
Ya Zhang\textsuperscript{\rm 1,3}
\quad 
Dacheng Tao\textsuperscript{\rm 4,2}
\\
\textsuperscript{\rm 1}Shanghai Jiaotong University, China \\
\textsuperscript{\rm 2}JD Explore Academy, China\\
\textsuperscript{\rm 3} Shanghai AI Laboratory, China\\
\textsuperscript{\rm 4} The University of Sydney, Australia\\
{\tt\small \{charles-hu,ya\_zhang\}@sjtu.edu.cn; \quad 
\{mathshenli,dacheng.tao\}@gmail.com
}
}


\maketitle
\ificcvfinal\thispagestyle{empty}\fi

\begin{abstract}

Offline reinforcement learning (RL) is a challenging task, whose objective is to learn policies from static trajectory data without interacting with the environment. Recently, offline RL has been viewed as a sequence modeling problem, where an agent generates a sequence of subsequent actions based on a set of static transition experiences. However, existing approaches that use transformers to attend to all tokens naively can overlook the dependencies between different tokens and limit long-term dependency learning. In this paper, we propose the Graph Decision Transformer (GDT), a novel offline RL approach that models the input sequence into a causal graph to capture potential dependencies between fundamentally different concepts and facilitate temporal and causal relationship learning. GDT uses a graph transformer to process the graph inputs with relation-enhanced mechanisms, and an optional sequence transformer to handle fine-grained spatial information in visual tasks. Our experiments show that GDT matches or surpasses the performance of state-of-the-art offline RL methods on image-based Atari and OpenAI Gym.
\end{abstract}

\begin{figure}
    \centering
    \includegraphics[width=0.45\textwidth]{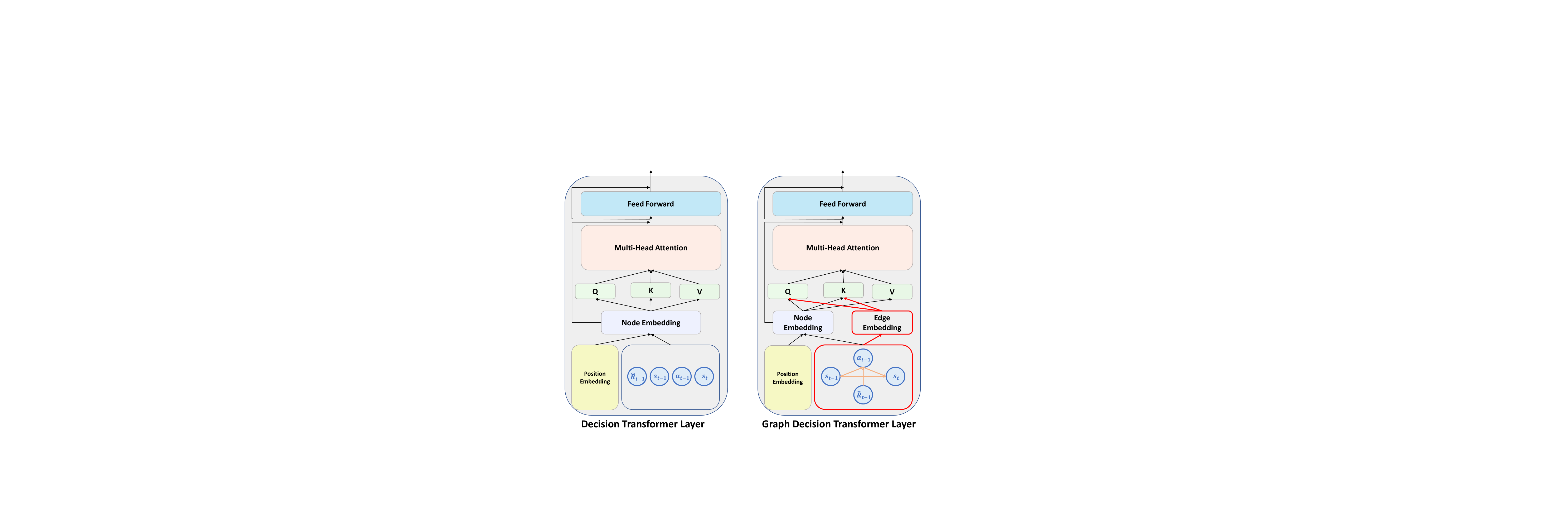}
    \caption{Comparison of GDT and DT. GDT employs additional edge embeddings and node embeddings to obtain Q and K, while using only node embeddings to obtain V. 
    %
    %
    }
    \label{fig:gtl}
\end{figure}

\section{Introduction}

Reinforcement Learning (RL) is inherently a sequential process where an agent observes a state from the environment, takes an action, observes the next state, and receives a reward. 
To model RL problems, Markov Decision Processes (MDPs) have been widely employed, where an action is taken solely based on the current state, which is assumed to encapsulate the entire history.
Online RL algorithms \cite{DQN, silver2017mastering} use the temporal difference (TD) learning to train agents by interacting with the environment, but this can be prohibitively expensive in real-world settings.
Offline RL \cite{levine2020offline}, on the other hand, seeks to overcome this limitation by learning policies from a pre-collected dataset, without the need to interact with the environment.
This approach makes RL training more practical for real-world scenarios and has therefore garnered significant attention.

Recent advances \cite{DT, TT} in offline RL have taken a new perspective on the problem, departing from conventional methods that concentrate on learning value functions \cite{riedmiller2005neural, kostrikov2021offline} or policy gradients \cite{precup2000eligibility, fujimoto2021minimalist}. 
Instead, the problem is viewed as a generic sequence modeling task, where past experiences consisting of state-action-reward triplets are input to Transformer \cite{attention}. 
The model generates a sequence of action predictions using a goal-conditioned policy, effectively converting offline RL to a supervised learning problem.
This approach relaxes the MDP assumption by considering multiple historical steps to predict an action, allowing the model to be capable of handling long sequences and avoid stability issues associated with bootstrapping \cite{srivastava2019training, kumar2019reward}. 
Furthermore, this framework unifies multiple components in offline RL, such as estimating the behavior policy and predictive dynamics modeling, into a single sequence model, resulting in superior performance.

However, this approach faces three major issues.
Firstly, states and actions represent fundamentally different concepts \cite{SPLT}. 
While the agent has complete control over its action sequences, the resulting state transitions are often influenced by external factors. 
Thus, modeling states and actions as a single sequence may indiscriminate the effects of the policy and world dynamics on the return, which can lead to overly optimistic behavior.
Secondly, in RL problems, the adjacent states, actions, and rewards are typically strongly connected due to their potential causal relationships. 
Specifically, the state observed at a given time step is a function of the previous state and action, and the action taken at that time step influences the subsequent state and reward. 
Simply applying Transformer to attend to all tokens without considering the underlying Markovian relationship can result in an overabundance of information, hindering the model's ability to accurately capture essential relation priors and handle long-term sequences of dependencies.
Finally, tokenizing image states as-a-whole using convolutional neural networks (CNNs) can hinder the ability of Transformers to gather fine-grained spatial relations. 
This loss of information can be especially critical in visual RL tasks that require detailed knowledge of regions-of-interest.
Therefore, it is necessary to find a more effective way to represent states and actions separately while still preserving their intrinsic relationship, and to incorporate the Markovian property and spatial relations in the modeling process.

To alleviate such issues, we propose a novel approach, namely Graph Decision transformer (GDT), which involves transforming the input sequence into a graph structure. 
The Graph Representation explicitly incorporates the potential dependencies between adjacent states, actions, and rewards, thereby better capturing the Markovian property of the input and differentiating the impact of different tokens.
To process the input graph, we utilize the Graph Transformer to effectively handle long-term dependencies that may be present in non-Markovian environments. 
%
To gather fine-grained spatial information, we incorporate an optional Sequence Transformer following StAR \cite{starformer} to encode image states as patches similar to ViT \cite{ViT}, which helps with action prediction and reduces the learning burden of the Graph Transformer.
%
Our experimental evaluations conducted in Atari and OpenAI Gym environments provide empirical evidence to support the advantages of utilizing a causal graph representation as input to the Graph Transformer in RL tasks. 
The proposed GDT method achieves state-of-the-art performance in several benchmark environments and outperforms existing offline RL methods.

In summary, our main contributions are four-fold:
\begin{itemize}
    \item We propose a novel approach named GDT, that transforms input sequences into graph structures to better capture potential dependencies between adjacent states, actions, and rewards and differentiate the impact of these different tokens.
    \item We utilize the Graph Transformer to process the input graph, which can effectively handle long-term dependencies that may be present in non-Markovian environments.
    \item We incorporate Sequence Transformer to encode image states as patches to gather fine-grained spatial information crucial for visual input environments.
    \item We extensively evaluate GDT on Atari and OpenAI Gym environments, demonstrating its superior performance compared to existing methods
\end{itemize}
%
%

\begin{figure*}[t]
    \centering
    \includegraphics[width=0.9\textwidth]{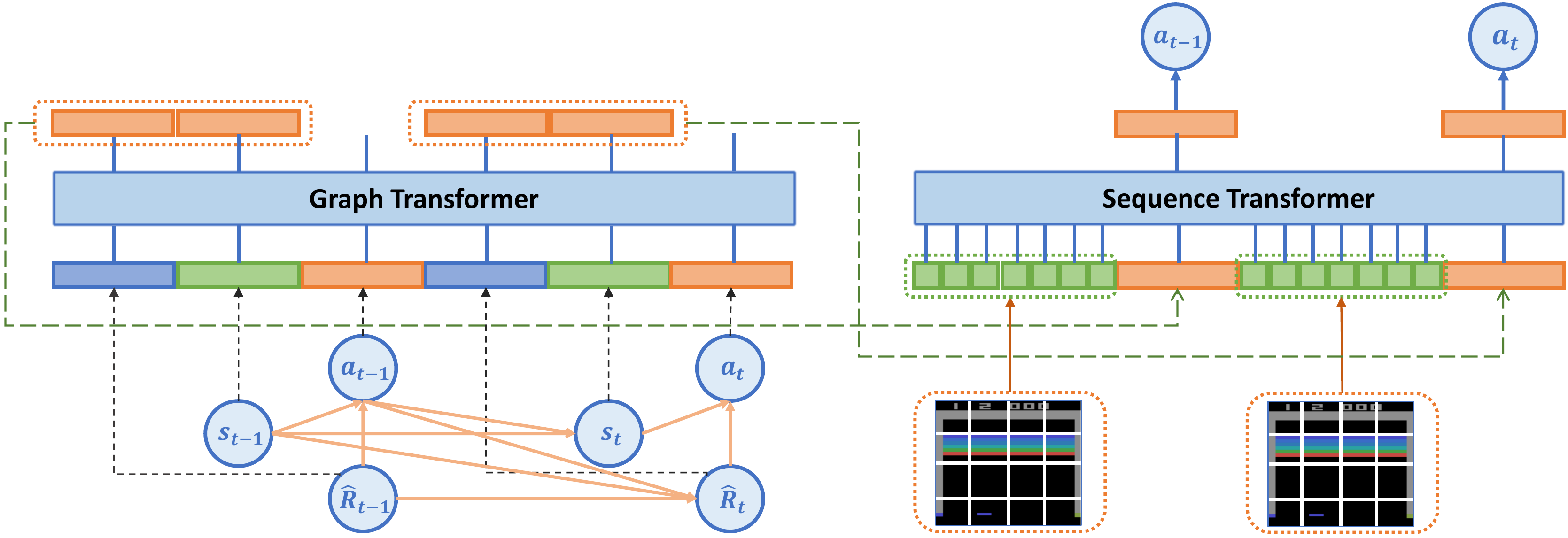}
    \caption{The proposed model comprises three main components: the Graph Representation, the Graph Transformer, and an optional Sequence Transformer. 
    When the output of the Graph Transformer is used directly for action prediction, the resulting model is referred to as GDT, which includes only the left half of the figure.
    On the other hand, if the output of the Graph Transformer is further processed by the Sequence Transformer, the resulting model is referred to as GDT-plus, which encompasses the entire figure.}
    \label{fig:model}
\end{figure*}

\begin{figure}
    \centering    \includegraphics[width=0.45\textwidth]{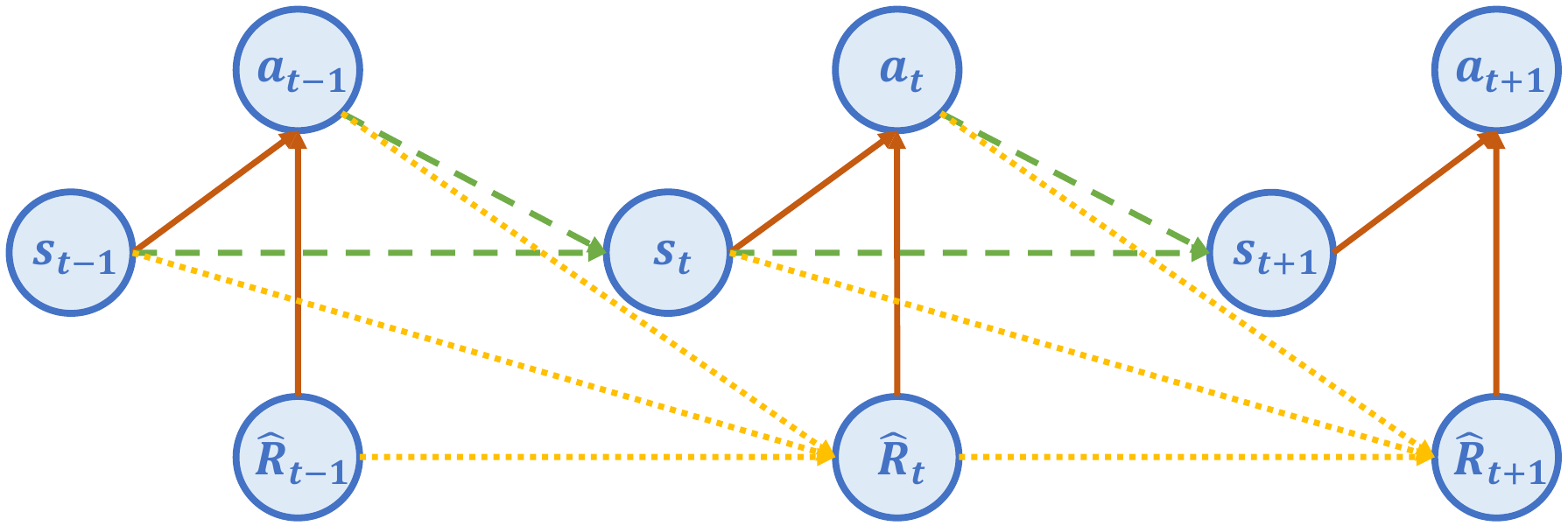}
    \caption{Input representation as a graph. 
    Nodes $\hat{R}_t$, $s_t$, and $a_t$ represent the return-to-go, state, and action at time $t$, respectively.  
    The directed edge connecting $s_{t-1}$ to $s_t$ indicates that the value of $s_{t-1}$ affects the value of $s_t$, and similar relationships exist for other directed edges.}
    \label{fig:input}
\end{figure}

\section{Related Work}

\paragraph{Offline RL}
Offline RL has recently gained significant attention as an alternative paradigm, where agents extract return-maximizing policies from fixed, limited datasets composed of trajectory rollouts from arbitrary policies \cite{levine2020offline}. 
These datasets, referred to as static datasets, are formally defined as $\mathcal{D}= \{(s_t, a_t, s_{t+1}, r_t)_i\}$, where $i$ is the index, the actions and states are generated by the behavior policy $(s_t, a_t) \sim d^{\pi_{\beta}}(\cdot)$, and the next states and rewards are determined by the dynamics $(s_{t+1}, r_t) \sim (T(\cdot | s_t, a_t), r(s_t, a_t))$.
Deploying off-policy RL algorithms directly in the offline setting is hindered by the distributional shift problem, which can result in a significant performance drop, as demonstrated in prior research \cite{fujimoto2019off}. 
To mitigate this issue, model-free algorithms aim to either constrain the action space of policy \cite{kumar2019stabilizing, siegel2020keep} or incorporate pessimism to value function \cite{CQL, fujimoto2019off}.
Conversely, model-based algorithms simulate the actual environment to generate more data for policy training \cite{kidambi2020morel, yu2020mopo}.
In this work, we propose a novel approach that avoids learning the dynamics model explicitly and directly generates the next action with the help of the Graph Transformer, enabling better generalization and transfer \cite{ramesh2021zero}.

\paragraph{RL to Sequence Modeling}
RL has recently garnered considerable interest as a sequence modeling task, particularly with the application of Transformer-based decision models \cite{TRL}.
The task is to predict a sequence of next actions given a sequence of recent experiences, including state-action-reward triplets. 
This approach can be trained in a supervised learning fashion, making it more amenable to offline RL and imitation learning settings.
Several studies \cite{bootstrapped, QDT, generalizedDT,zhang2023saformer} have explored the use of Transformers in RL under the sequence modeling pattern. 
For example, Chen et al. \cite{DT} train a Transformer as a model-free context-conditioned policy, while Janner et al. \cite{TT} bring out the capability of the sequence model by predicting states, actions, and rewards and employing beam search. 
Zheng et al. \cite{ODT} further fine-tune the Transformer by adapting this formulation to online settings.
Shang et al. \cite{starformer} explicitly model StAR-representations to introduce a Markovian-like inductive bias to improve long-term modeling.
In this work, we propose a graph sequence modeling approach to RL, which explicitly introduces the Markovian property to the representations. 
Our proposed GDT method outperforms several state-of-the-art non-Transformer offline RL and imitation learning algorithms on Atari and Gym benchmarks, demonstrating the advantage of incorporating graph structures in sequence modeling for RL tasks.

\paragraph{RL with Graph}
In recent years, the integration of graph neural networks (GNNs) with RL has attracted considerable attention for graph-structured environments \cite{munikoti2022challenges}. 
Specifically, GNNs can be combined with RL to address sequential decision-making problems on graphs. 
Existing research primarily focuses on using deep RL to improve GNNs for diverse purposes, such as neural architecture search (NAS) \cite{autognn}, enhancing the interpretability of GNN predictions \cite{shan2021reinforcement}, and designing adversarial examples for GNNs \cite{RLS2V, sun2020non}. 
Alternatively, GNNs can be utilized to solve relational RL problems, such as those involving different agents in a multi-agent deep RL (MADRL) framework \cite{graphcomm, DCG, zhang2021structural}, and different tasks in a multi-task deep RL (MTDRL) framework \cite{nervenet,huang2020one, battaglia2018relational,huang2022curriculum}. 
Despite the growing interest in this field, there is currently a lack of research on utilizing a Markovian dependency graph as input to GNNs for action prediction. 
Such an approach has strong potential due to the causal relationship between the constructed graph and its ability to be employed in various RL environments. 
This article will explore this approach in detail, offering a novel contribution to the field. 
Additionally, we will highlight the advantages of using GDT for action prediction in comparison to existing state-of-the-art algorithms.
\section{Methodology}
The proposed approach is a deep learning-based model for sequence modeling in RL tasks, which combines the advantages of graph and sequence modeling. 
The model consists of three main components: the Graph Representation, the Graph Transformer, and an optional Sequence Transformer, as shown in Figure \ref{fig:model}. 
Specifically,  Graph Representation is used to represent the input sequence as a graph with a causal relationship, which captures the dependencies between fundamentally different concepts.
The Graph Transformer then processes the graph inputs using relation-enhanced mechanisms, which allows the model to acquire long-term dependencies and model the interactions between different time steps of the graph tokens given reasonable causal relationships.
The optional Sequence Transformer is introduced to gather the fine-grained spatial information in the input, which is particularly important in visual tasks such as Atari environment.

The proposed approach offers several advantages. 
Firstly, it can effectively acquire the intricate dependencies and interactions between different time steps in the input sequence, making it well-suited for RL tasks. 
Secondly, it encodes the sequence as a causal graph, which explicitly incorporates potential dependencies between adjacent tokens, thereby facilitating the capture of the Markovian property and differentiating the impact of different tokens.
Finally, it can accurately gather fine-grained spatial information and integrate it into action prediction, leading to improved performance.
In summary, our proposed approach offers a powerful and flexible solution for RL tasks. 
Its unique design principles, which combine graph and sequence modeling, enable it to effectively obtain the dependencies and interactions between different time steps and tokens, making it a promising approach for various applications.
In the following sections, we provide a detailed description of each component of our proposed approach, outlining its input and output and discussing its underlying design principles.
 
\subsection{Graph Representation}  
Our approach constructs the input graph from trajectory elements based on their dependencies, capturing the temporal and causal relationships of the data. 
Each trajectory element represents a node in the graph, and edges are added accordingly. 
This approach considers the Markovian nature of the data and differs from the DT \cite{DT} serialized input.
Specifically, our approach considers the inputs to the trajectory as returns-to-go $\hat{R}_t = \sum_{t'=t}^T r_{t'}$, states $s_t$, and actions $a_t$, as illustrated in Figure \ref{fig:input}. 
Action $a_t$ selection depends on the current state $s_t$ and return-to-go $\hat{R}_t$. 
State $s_t$ generation is dependent on the previous state $s_{t-1}$ and action $a_{t-1}$. 
The return-to-go $\hat{R}_t$ is jointly determined by the previous return-to-go $\hat{R}_{t-1}$, state $s_{t-1}$, and action $a_{t-1}$.

The MDP is a framework where an agent is asked to make a decision based on the current state $s_t$; then, the environment responds to the action made by the agent and transitions the state to the next state $s_{t+1}$ with a reward $r_t$. 
We model these dependencies as directed edges and exclude the agent and environment objects to enable the Graph Transformer to better analyze the causal relationships between different tokens.
Through this approach, we explicitly embed the Markovian relationship into the model and distinguish the effect of distinct tokens, avoiding the overabundance of information that may prevent the model from accurately capturing essential relation priors.

\begin{figure*}[!t]
    \centering
    \includegraphics[width=6in]{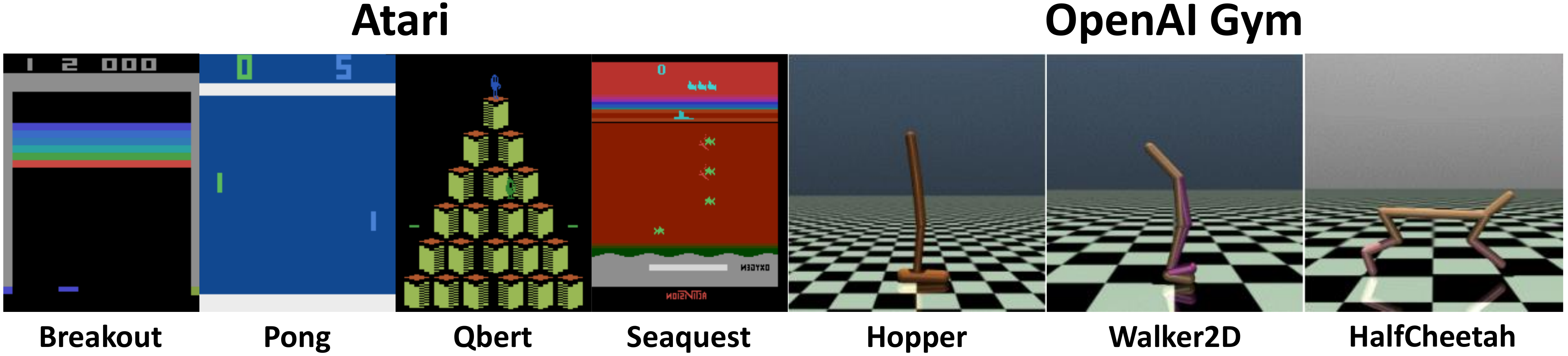}
    \caption{The environments utilized in this study include four Atari environments on the left, which accept images as inputs and have a discrete action space, and three OpenAI Gym environments on the right, which accept vectors as inputs and have a continuous action space.}
    \label{fig:env}
\end{figure*}

\subsection{Graph Transformer} 

Our approach employs autoregressive modeling to predict current action using the graph representation as input to the Graph Transformer. 
The Graph Transformer plays a central role in extracting hidden features from the graph. 
In some scenarios, such as Atari games, past information also plays a critical role, thus we employ a global attention mechanism, allowing each node to not only observe its directly connected nodes but also all nodes preceding the current moment.
In the vanilla multi-head attention, the attention score between the element $x_i$ and $x_j$ can be formulated as the dot-product between their query vector and key vector, respectively:
\begin{equation}
\small
    s_{ij} = f(x_i, x_j) = x_i W_q^T W_k x_j.
\end{equation}
In the case of Graph Transformers, the computation of attention scores also needs to account for the temporal and causal relationships. 
To achieve this, we follow the concept of relation-enhanced mechanism proposed in \cite{cai2020graph}, which utilizes both node and relation representations in computing attention scores, as shown in Figure \ref{fig:gtl}.
The equation for the computation of attention scores is shown below:
\begin{equation}
\small
    \begin{split}
        s_{ij} &= g(x_i, x_j, r_{ij}) \\
            &= (x_i + r_{i \rightarrow j})W_q^TW_k(x_j + r_{j \rightarrow i}),
    \end{split}
\end{equation}
where $j \leq i$ and $r_{ij}$ is learned through an embedding layer that takes the adjacency matrix as input.
The incorporation of relation representation enables the model to take into account the plausible causal relationships, which relieves the burden on the Graph Transformer of learning potential long-term dependencies among node vectors at different time steps. 
This observation has been empirically validated through our ablation experiments.

The input to the $l$-th layer of the Graph Transformer is a token graph. 
To preserve temporal relationships, the graph is pruned to form the following sequence $\overline{G}$ ($\overline{\cdot}$ means pruning into sequence):
\begin{equation}
\small
    \begin{split}
        \overline{G}_{in}^l &= \{\hat{R}_0^{l-1}, s_0^{l-1}, a_0^{l-1},  \dots, \hat{R}_T^{l-1}, s_T^{l-1}, a_T^{l-1}  \}. \\
    \end{split}
\end{equation}
Each token graph is transformed to $g_t^l$ by a Graph Transformer layer:
\begin{equation}
\small
    \begin{split}
        G_{out}^l &=  F^l_{graph} (G_{in}^l) = F^{l}_{graph} (G_{out}^{l-1}), \\
        g_t^l :&= \text{FC} (\overline{G}_{out}^l [ 1 + 3t] , \overline{G}_{out}^l [ 3t]).
    \end{split}
\end{equation}
As shown in Figure \ref{fig:input}, the action $a_t$ is determined by both $\hat{R}_t$ and $s_t$. 
Thus, the feature vector $g_t^l$ is obtained by concatenating the two inputs and fed into a fully connected layer (with indexing starting from 0).
The feature $g_t^l$ extracted from the $l$-th layer of the Graph Transformer can be directly used to predict the action $\hat{a}_t = \phi(g_t^l)$, or be further processed by the subsequent Sequence Transformer.

\subsection{Sequence Transformer} 
Motivated by  \cite{starformer}, we introduce the Sequence Transformer to assist with action prediction and reduce the learning burden of the Graph Transformer. 
The Sequence Transformer adopts the conventional Transformer layer design from \cite{attention} and is incorporated to gather fine-grained spatial information that is crucial for visual input environments.
The initial layer of the Sequence Transformer takes a collection of image patches and $g_t^0$ as inputs: 
\begin{equation}
\small
    Y_{in, t}^0 = \{z_{s_{t}^1}, z_{s_{t}^2}, \dots,  z_{s_{t}^n}, g_t^0 \},
\end{equation}
where $n$ is the number of image patches, and the feature vector $g_t$ is positioned after state patches $\{ z_{s_t^i} \}_{i=1}^n$, which enables $g_t$ to attend to all spatial information.
We have $T$ groups of such token representations, which are simultaneously processed by the Sequence Transformer:
\begin{equation}
\small
    \begin{split}
        Y_{in}^0 &= \sum_{t=0}^T Y_{in, t}^0, \\
        Y_{out}^0 &= F_{sequence}^0(Y_{in}^0), \\
    \end{split}
\end{equation}
where adding means concatenating the contents of two collections.
The subsequent layer of the Sequence Transformer takes the fusion of the previous layer's output and $g_t^l$ as its input.
This is achieved by adding $g_t^l$ to the position that corresponds to $g_t^{l-1}$ in the output sequence while leaving the other parts of the output sequence unchanged.
The formulation for this operation is as follows:
\begin{equation}
\small
\!\! Y_{in}^l[i] \!=\! \hat{Y}_{out}^{l-1}[i] \!=\! 
        \begin{cases}
            Y_{out}^{l-1}[i] + g_t^l, & i = n + t(n+1), \\
                                & t=0,1,2,\dots,T \\
            Y_{out}^{l-1} [i], & \text{otherwise}
        \end{cases}
\end{equation}
The output feature $h_t^l:= Y_{out}^l[ n + t(n+1)]$ extracted from the $l$-th layer of the Sequence Transformer is fed into a linear head to predict the action, denoted as $\hat{a}_t = \phi(h_t^l)$ when it is the final layer.
Further elaboration on this connection method will be provided in Sec. \ref{subsec:ablation}.

\subsection{Training procedure}

GDT is a drop-in replacement for DT as the training and inference procedures remain the same. 
However, additional graph construction for the input sequence is required for GDT. 
Specifically, a graph $G=(X,A)$ is constructed from a minibatch of length $K$ (total $3K$ tokens: return-to-go, state, and action)
where $X$ represents the node embedding matrix, and $A$ represents the edge embedding matrix.
The constructed graph is then input into the Graph Transformer, and attention scores are calculated using both $X$ and $A$. 
%
The learning objective for discrete environments can be formulated as follows:
\begin{equation}
\small
  \!\!\!  \mathbb{E}_{(\hat{R}, s, a) \sim \mathcal{T}} \left[\! \frac{1}{T} \sum_{t=1}^T (a_t \!-\! \pi_{\text{GDT}}(s_{-K: t}, \hat{R}_{-K: t}, a_{-K:t-1}) )^2 \right].\!
\end{equation}


To end this section, we give several comments on the proposed GDT method. 
Compared with DT, which uses a serialized input, GDT represents the input sequence as a graph with a causal relationship, enabling the Graph Transformer to capture dependencies between fundamentally different concepts. 
On the other hand, compared with methods that implicitly learn Markovian patterns \cite{TT} or introduce an additional step Transformer \cite{starformer}, GDT directly incorporates the Markovian relationship in the input. 
This feature allows the model to handle dependencies between sequences and tokens effectively, leading to improved performance without additional computational overhead. 
Additionally, the introduced Sequence Transformer can maintain fine-grained spatial information using ViT-like image patches, which is particularly important in visual tasks and can improve action prediction accuracy in such environments.
Although GDT's performance in vector state environments is not significantly improved compared to that in visual environments, it is still effective in such relatively simple environments (Sec. \ref{subsec:gym}).

\begin{figure*}[!t]
    \centering
    \includegraphics[width=6in]{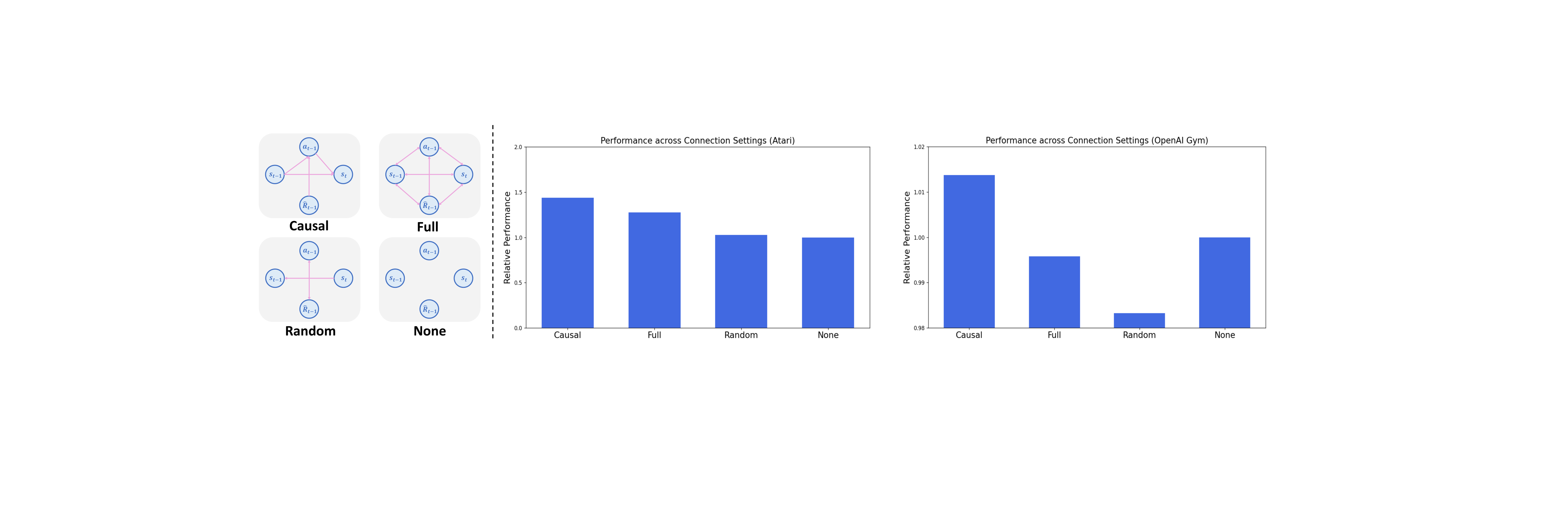}
    \caption{
    Comparison of different graph connection methods on Atari and OpenAI Gym Environments (same in later experiments). The left panel illustrates four different graph connection methods: Causal, Full, Random, and None. The right panel shows the normalized performance comparison of these methods.}
    \label{fig:graph}
\end{figure*}

\begin{figure*}[!t]
    \centering
    \includegraphics[width=6in]{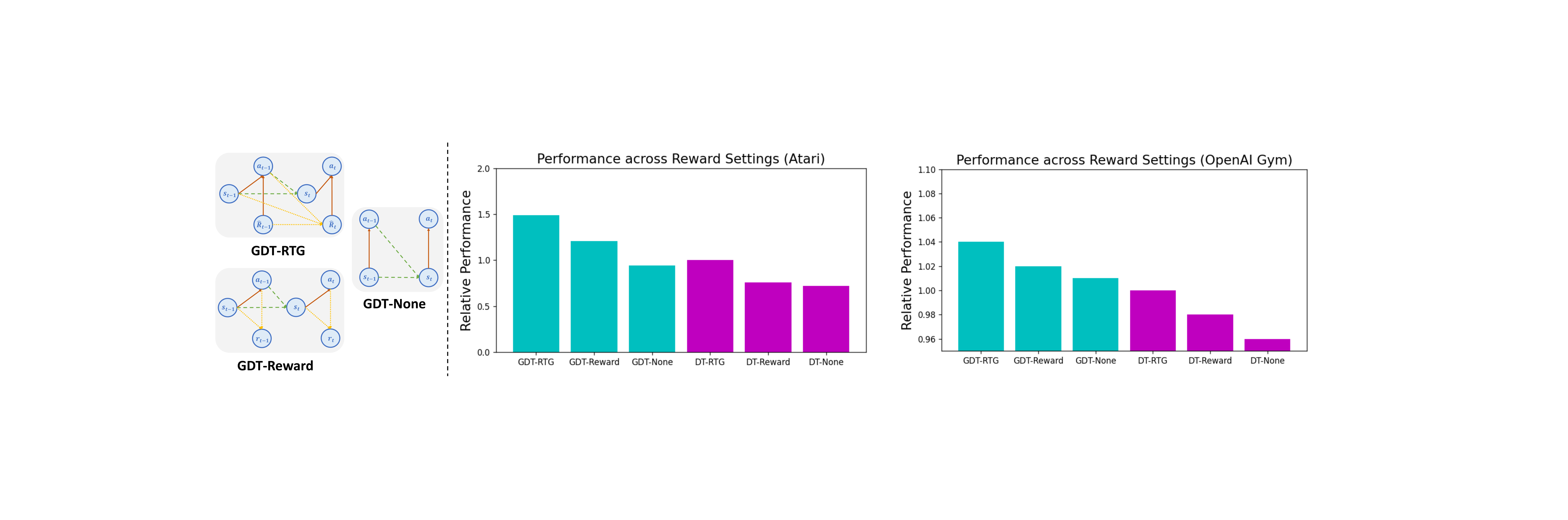}
    \caption{
    The left panel depicts the input graph structure of the GDT for three reward settings: return-to-go (RTG), stepwise reward (Reward), and no reward (None). The right panel displays the performance comparison between GDT and DT}
    \label{fig:reward}
\end{figure*}

\section{Experiment}

In this section, we provide a comprehensive evaluation of the proposed Graph Decision Transformer (GDT) model, which is designed to capture the complex relationships among graph-structured data and make effective decisions based on them. 
Our main objective is to assess the effectiveness of GDT in comparison to two popular algorithms: offline algorithms based on TD-learning and imitation learning algorithms.
TD-learning is a widely adopted algorithm in RL due to its remarkable sampling efficiency and impressive performance on various RL tasks. 
It has become a benchmark algorithm for many RL applications. 
On the other hand, imitation learning algorithms, represented by DT, have gained increasing attention in recent years due to their ability to learn from expert demonstrations and achieve performance comparable to TD-learning in various RL tasks.
%
%
We conducted a comprehensive evaluation of the performance of the GDT model on a range of tasks. 
Specifically, just as Figure \ref{fig:env} shown, we evaluated the performance of GDT on the widely used Atari benchmark \cite{atari}, which consists of a set of discrete control tasks, as well as on the OpenAI Gym \cite{gym}, which comprises a variety of continuous control tasks.


\subsection{Atari}

\begin{table*}[]
\centering
\caption{Results for 1\% DQN-replay Atari datasets. We evaluate the performance of GDT on four Atari games using three different seeds, and report the mean and variance of the results. 
Best mean scores are highlighted in bold.
The assessment reveals that GDT surpasses conventional RL algorithms on most tasks and achieves better performance than DT across all games. 
In contrast, GDT-plus achieves comparable results to StAR, leveraging fine-grained spatial information.}
\label{tab:atari}
\vspace{.2cm}
\scalebox{0.85}{%
\begin{tabular}{>{\centering}p{0.10\textwidth}>{\centering}p{0.06\textwidth}>{\centering}p{0.08\textwidth}>{\centering}p{0.06\textwidth}>{\centering}p{0.12\textwidth}>{\centering}p{0.12\textwidth}>{\centering}p{0.12\textwidth}>{\centering}p{0.12\textwidth}>{\centering\arraybackslash}p{0.12\textwidth}}
\toprule
Game     & CQL   & QR-DQN & REM & BC           & DT           & StAR         & GDT & GDT-plus \\ \midrule
Breakout & 211.1 & 17.1   & 8.9 & 138.9 $\pm$ 61.7 & 267.5 $\pm$ 97.5 & 436.1 $\pm$ 63.6 &    393.5 $\pm$ 98.8   &   \textbf{441.7 $\pm$ 41.0 }   \\
Qbert    & \textbf{ 104.2 }& 0      & 0   & 17.3 $\pm$ 14.7  & 15.4 $\pm$ 11.4  & 51.2 $\pm$ 11.5  &    45.5 $\pm$ 14.6    &   51.7 $\pm$ 20.8    \\
Pong     & \textbf{ 111.9} & 18     & 0.5 & 85.2 $\pm$ 20.0  & 106.1 $\pm$ 8.1  & 110.8 $\pm$ 60.3 &  108.4 $\pm$ 4.7     &   111.2 $\pm$ 0.9              \\
Seaquest & 1.7   & 0.4    & 0.7 & 2.1 $\pm$ 0.3    & 2.5 $\pm$ 0.4    & 1.7 $\pm$ 0.3    & \textbf{  2.8 $\pm$ 0.1 }      &     2.7 $\pm$ 0.1   \\ \bottomrule
\end{tabular}%
}
\end{table*}

The Atari benchmark \cite{atari} is a well-recognized and widely-adopted benchmark for evaluating the performance of RL algorithms. 
%
%
In this study, we choose four games from the Atari benchmark, namely Breakout, Pong, Qbert, and Seaquest, each of which requires the agent to handle high-dimensional visual inputs and complex credit assignment.
Similar to prior work \cite{DT}, we construct the offline dataset by sampling 1\% of the DQN replay buffer dataset \cite{REM}, which consists of nearly 500k transition steps.
To enable fair comparisons, we follow the normalization protocol proposed in \cite{hafner2020mastering}, where the final scores are normalized such that a score of 100 represents the expert level performance and a score of 0 represents the performance of a random policy.

For baseline benchmark, we compare GDT with TD-learning based algorithms, including CQL \cite{CQL}, QR-DQN \cite{QRDQN}, and REM \cite{REM}, and several imitation learning algorithms, including DT \cite{DT}, StARformer \cite{starformer}, and behavior cloning (BC), and report the results from original papers. 
%

Table \ref{tab:atari} presents the comparison of our proposed method with CQL, REM, QR-DQN, BC, and DT methods on four games. The results show that our method achieves comparable performance to CQL in three out of four games, while significantly outperforming the other methods in all four games. 
This indicates that our approach, which introduces the causal relationship in the input and leverages the Graph Transformer accordingly, is superior to the other methods.
To ensure a fair comparison with StAR, we further introduce a Sequence Transformer to incorporate fine-grained spatial information and report the results as GDT-plus. 
%
The results demonstrate that GDT-plus achieves comparable or superior performance to StAR on all four Atari games, emphasizing the significance of fine-grained information in the games.
Compared with GDT, the success of the Sequence Transformer in incorporating such information into action prediction is also highlighted.

\subsection{OpenAI Gym}
\label{subsec:gym}

\begin{table*}[tb!]
\centering
\caption{Results for D4RL datasets. The performance of GDT is evaluated using three different seeds, and the mean and variance are reported. 
Best mean scores are highlighted in bold.
The results demonstrate that GDT exhibits superior performance compared to conventional RL algorithms on most of the evaluated tasks, and GDT-plus performs comparably to GDT.}
\label{tab:d4rl}
\vspace{.2cm}
\scalebox{0.75}{%
\begin{tabular}{>{\centering}p{0.14\textwidth}>{\centering}p{0.10\textwidth}>{\centering}p{0.06\textwidth}>{\centering}p{0.06\textwidth}>{\centering}p{0.08\textwidth}>{\centering}p{0.08\textwidth}>{\centering}p{0.06\textwidth}>{\centering}p{0.06\textwidth}>{\centering}p{0.10\textwidth}>{\centering}p{0.10\textwidth}>{\centering}p{0.10\textwidth}>{\centering\arraybackslash}p{0.10\textwidth}}
\toprule
Dataset & Environment & CQL & BEAR & BRAC-v & AWR & MBOP & BC & DT & StAR & GDT & GDT-plus \\ \midrule
Medium-Expert & HalfCheetah  & 62.4 & 53.4 & 41.9 & 52.7 &\textbf{ 105.9} & 59.9 & 86.8 $\pm$ 1.3 & 93.7 $\pm$ 0.1  & 92.4 $\pm$ 0.1   & 93.2 $\pm$ 0.1 \\
Medium-Expert & Hopper       & 111.0  & 96.3 & 0.8  & 27.1 & 55.1  & 79.6 & 107.6 $\pm$ 1.8 & \textbf{111.1 $\pm$ 0.2} & 110.9 $\pm$ 0.1  & \textbf{111.1 $\pm$ 0.1} \\
Medium-Expert & Walker       & 98.7 & 40.1 & 81.6 & 53.8 & 70.2  & 36.6 & 108.1 $\pm$ 0.2 & 109.0 $\pm$ 0.1 & \textbf{109.3 $\pm$ 0.1}  &  107.7 $\pm$ 0.1\\ \midrule
Medium        & HalfCheetah  & 44.4 & 41.7 & \textbf{46.3} & 37.4 & 44.6  & 43.1 & 42.6 $\pm$ 0.1 & 42.9 $\pm$ 0.1  &  42.9 $\pm$ 0.1 &  42.9 $\pm$ 0.1 \\
Medium        & Hopper       & 58.0   & 52.1 & 31.1 & 35.9 & 48.8  & 63.9 & 67.6 $\pm$ 1.0 & 59.5 $\pm$ 4.2  & 65.8 $\pm$ 5.8  & \textbf{77.1 $\pm$ 2.5} \\
Medium        & Walker       & 79.2 & 59.1 & \textbf{81.1} & 17.4 & 41.0    & 77.3 & 74.0 $\pm$ 1.4 & 73.8 $\pm$ 3.5 &  77.8 $\pm$ 0.4 & 76.5 $\pm$ 0.7 \\ \midrule
Medium-Replay & HalfCheetah  & 46.2 & 38.6 & \textbf{47.7} & 40.3 & 42.3  & 4.3  & 36.6 $\pm$ 0.8 & 36.8 $\pm$ 3.3  & 39.9 $\pm$ 0.1 & 40.5 $\pm$ 0.1 \\
Medium-Replay & Hopper       & 48.6 & 33.7 & 0.6  & 28.4 & 12.4  & 27.6 & 82.7 $\pm$ 7.0 & 29.2 $\pm$ 4.3 & 81.6 $\pm$ 0.6  & \textbf{85.3 $\pm$ 25.2}  \\
Medium-Replay & Walker       & 26.7 & 19.2 & 0.9  & 15.5 & 9.7   & 36.9 & 66.6 $\pm$ 3.0 & 39.8 $\pm$ 5.1  & 74.8 $\pm$ 1.9 & \textbf{77.5 $\pm$ 1.3}  \\ \midrule
\multicolumn{2}{c}{\textbf{Average}} & 63.9 & 48.2 & 36.9 & 34.3 & 47.8  & 46.4 & 74.7 & 66.2 & 76.8 & \textbf{79.1} \\ \bottomrule
\end{tabular}%
}
\end{table*}

The D4RL benchmark \cite{d4rl} evaluates the performance of RL algorithms in continuous control tasks, particularly in robotic manipulation tasks with challenging control and decision-making in continuous action spaces. 
%
%
%
In this study, we select three games from the standard locomotion environments in Gym, namely HalfCheetah, Hopper, and Walker, and three different dataset settings, namely medium, medium-replay, and medium-expert.
To ensure fair comparisons, we also normalize the scores according to the protocol established in \cite{d4rl}, where a score of 100 corresponds to an expert policy. 
%
%

For baseline benchmark, we compare GDT with a range of model-free methods, including CQL \cite{CQL}, BEAR \cite{BEAR}, BRAC \cite{BRAC}, AWR \cite{AWR}, and model-based methods, including MBOP \cite{MBOP}, a current strong baseline with model-based offline RL algorithms. 
We also compare the performance of GDT with several imitation learning algorithms, including DT \cite{DT}, StARformer \cite{starformer}, and BC. 
The performance of CQL, MBOP, and DT are reported from the original papers, while the results of BEAR, BRAC, and AWR are reported from the D4RL paper, and the other methods are run by us for a better fair comparison. 
This comprehensive evaluation encompasses a variety of techniques and conducts a thorough examination of the effectiveness of GDT in comparison to the existing state-of-the-art algorithms.

The results presented in Table \ref{tab:d4rl} demonstrate the superior performance of GDT on most of the evaluated tasks, with comparable performance to the state-of-the-art algorithm on the remaining tasks.
In the Gym environment, the input state is represented as a vector rather than an image, as is the case in the Atari environment. 
Therefore, when introducing the Sequence Transformer, the fine-grained spatial information is still obtained by embedding the state vector. 
In this context, the performance of GDT-plus has not shown significant improvement compared to GDT, and in some cases even achieves negative effects, which is also reflected in the results of StAR.
This may be attributed to the increased complexity of the model due to the Sequence Transformer, leading to overfitting on the state vector and producing a counterproductive effect.
\begin{figure}[!t]
    \centering
    \includegraphics[width=3in]{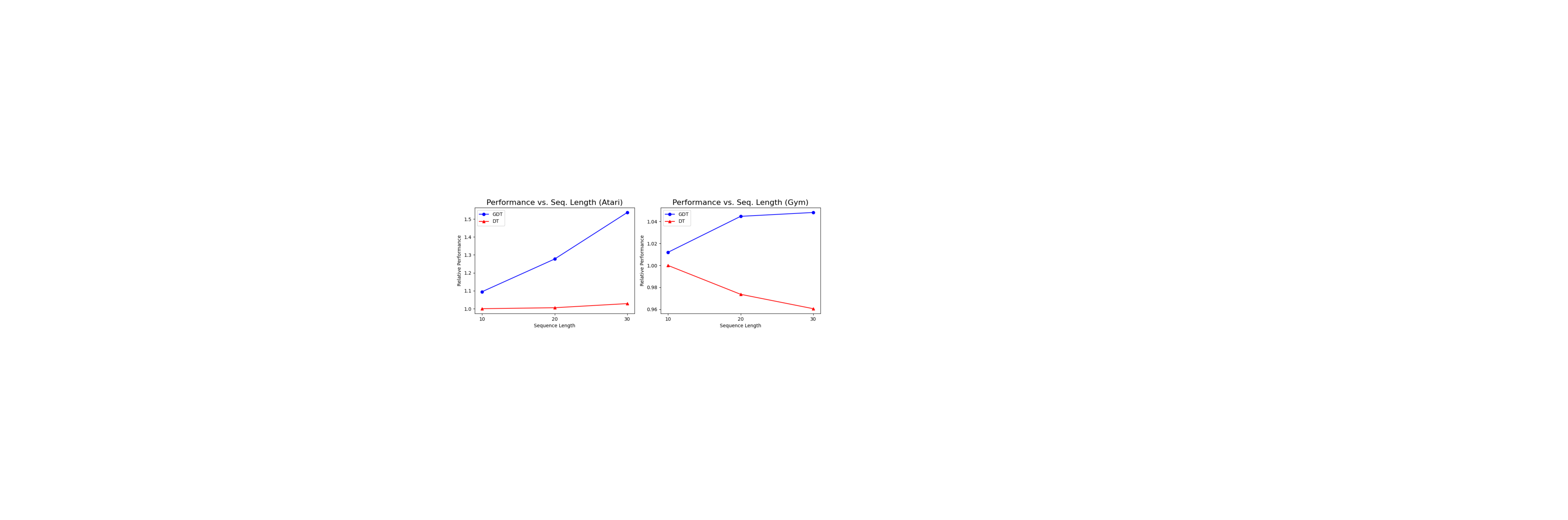}
    \caption{The performance comparison of different input lengths $T \in \{10,20,30 \}$.}
    \label{fig:length}
\end{figure}

\begin{figure*}[!t]
    \centering
    \includegraphics[width=6in]{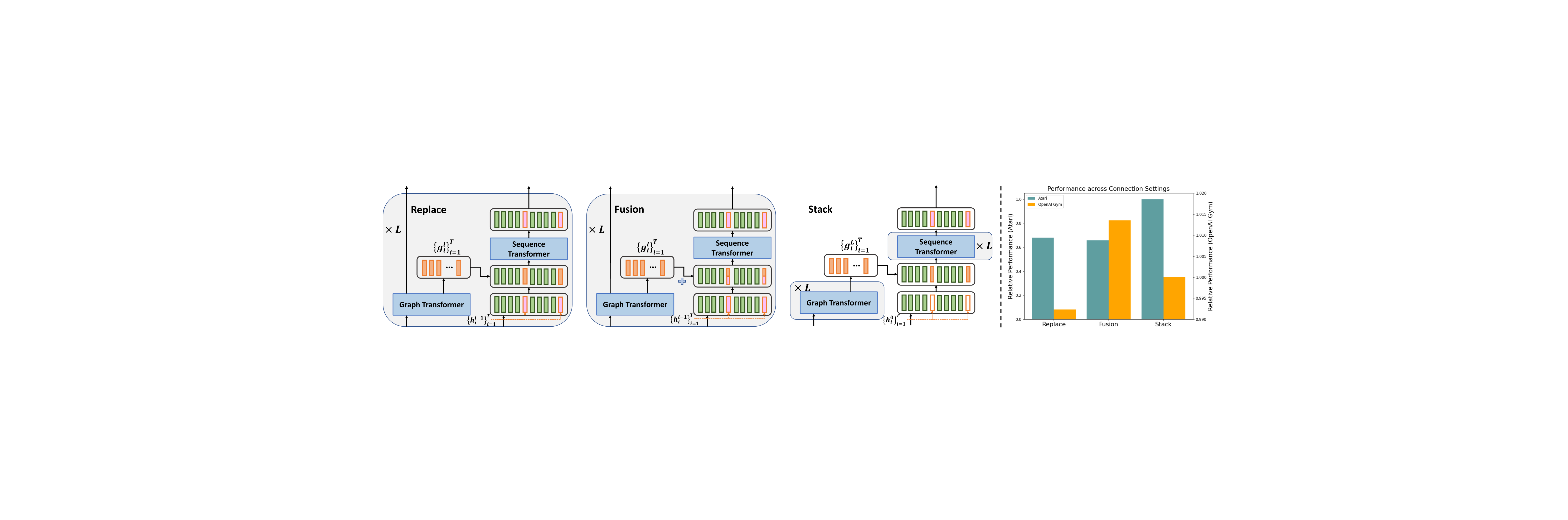}
    \caption{Performance comparison of Sequence Transformer with different connection methods. 
    The left panel illustrates the three connection methods, namely Replace, Fusion, and Stack. 
    The right panel shows the corresponding performance comparison.}
    \label{fig:connection}
\end{figure*}

\subsection{Ablations}
\label{subsec:ablation}

In comparison to the DT model, GDT transforms the input sequence into a graph with a causal relationship and feeds it into the Graph Transformer. 
Therefore, we initially investigate the impact of the graph structure on the performance of GDT. 
Additionally, we examine the influence of reward settings, which have a greater effect on the performance of the DT model. 
Lastly, we conduct ablation experiments on the length of the input sequence to examine the Graph Transformer's ability to rely on long sequences. 
To capture fine-grained spatial information, we introduce an additional Sequence Transformer to refine the action prediction process. 
We explore three ways to connect the Sequence Transformer. 
To provide a comprehensive evaluation, all ablation experiments will be conducted on both Atari and OpenAI Gym (Medium) environments.

\textbf{Graph representation setting.}
%
As we transform the input sequence into a graph with causal relationships, we aim to investigate the impact of the graph structure on the overall performance. 
Given that the graph connection method has a complexity of $\mathcal{O}(n^2)$, we primarily explore four different connection methods: causal connection, full connection, random connection, and none connection. 
%
%
The results in Figure \ref{fig:graph} demonstrate that the causal connection method outperforms the other methods in both environments. 
The full and none connection methods yield similar performances, while the random Connection method is significantly impacted by the environment. 
Although this connection method can still achieve better results when historical information is crucial, it can be counterproductive in environments that are more reliant on causal relationships.

\textbf{Reward setting.}
%
In this study, we investigate the impact of reward setting on GDT's performance. 
Specifically, we examine the effect of three reward settings, namely return-to-go (RTG), stepwise reward (Reward), and none reward (None). 
RTG is a commonly used setting in sequence modeling methods \cite{eysenbach2020rewriting, li2020generalized, srivastava2019training}, and its experimental results are greatly influenced by the value of the return-to-go hyper-parameter.
Reward refers to the immediate reward generated by the environment after each step, which is commonly used in traditional TD-learning-based algorithms \cite{hessel2018rainbow, PPO, SAC}.
The None setting usually corresponds to straightforward BC.

The results are presented in Figure \ref{fig:reward}. 
It should be noted that since GDT takes a graph with a causal relationship as input, the causal relationship changes correspondingly for each reward setting, as shown in the left part of the figure. 
The introduction of rewards has improved the performance of both methods compared to the no reward setting. 
However, DT is more reliant on return-to-go than GDT, and there is a large difference in performance between the two reward settings for DT. 
Importantly, GDT still outperforms DT with the return-to-go setting under the stepwise reward setting.
This demonstrates that introducing causality in the input can reduce the dependence of sequence modeling on return-to-go.
Furthermore, under the stepwise reward setting, GDT's graph input form is similar to the dynamic modeling method in model-based algorithms, which is worth further research to expand and introduce in model-based approaches.

\textbf{Trajectory input length.}
%
In this experiment, we investigate the effect of input sequence length on the performance of GDT. 
While in a Markovian environment, the state at the previous moment is often sufficient to determine the current action, the DT experiment \cite{DT} reveals that past information is valuable for the sequence modeling method in Atari environments, where longer sequences tend to yield better results than those of length $1$.
Subsequently, we explore the impact of different sequence lengths on performance and compare the results of DT and GDT, demonstrating the superior ability of GDT to handle long sequence inputs.

Figure \ref{fig:length} presents the impact of sequence length on DT and GDT performance.
It is observed that DT's performance remains relatively unchanged as the sequence length increases, while GDT exhibits a significant improvement, demonstrating its superior ability to handle long sequences. 
By incorporating causal relationships in the input, GDT enables the Graph Transformer to effectively handle the dependencies between long sequences, leading to improved performance without additional computational overhead. 
%

\textbf{Sequence Transformer connection method.}
%
To capture fine-grained spatial information, we introduce an optional Sequence Transformer to improve action prediction. 
We denote the output variables of Graph Transformer as $g_t^l$ and the corresponding input variables in Sequence Transformer as $h_t^l$. There are three possible ways to connect the two: (1) in each layer, $g_t^l$ replaces $h_t^{l-1}$ (GDT-Replace); (2) in each layer, $g_t^{l}$ is added to $h_t^{l-1}$ (GDT-Fusion); and (3) the last layer of $g_t^{L}$ is used as the initial layer $h_t^0$ (GDT-Stack).
%
The experimental results in Figure \ref{fig:connection} demonstrate that in the Atari environment with high information density, stacking features at the end is the most effective approach for feature refinement. 
In contrast, for the Gym environment with less information, feature refinement through the fusion of different abstraction levels achieves better performance.

\section{Conclusion}

In conclusion, this paper proposes the Graph Decision Transformer (GDT), a novel offline RL approach that models input sequences into causal graphs to capture potential dependencies between fundamentally different concepts and facilitate temporal and causal relationships learning. 
We show that GDT outperforms state-of-the-art offline RL methods on image-based Atari and OpenAI Gym.


Our work highlights the potential of graph-structured inputs in RL, which has been less explored compared to other deep learning domains. 
 We believe that our proposed GDT approach can inspire further research in RL, especially in tasks where spatial and temporal dependencies are essential, such as robotics, autonomous driving, and video games.
 Further exploration of graph-structured inputs may lead to more efficient and effective RL algorithms with broader applicability in real-world scenarios.

{\small
\bibliographystyle{ieee_fullname}
\bibliography{egbib}
}

\clearpage
\appendix
\noindent{\Large \textbf{Appendix}}

\section{Experimental Details}


\subsection{Baseline} 
For the Atari environment, we evaluate GDT against several state-of-the-art non-Transformer offline RL methods, such as CQL\cite{CQL}, QR-DQN\cite{QRDQN}, and REM \cite{REM}, and several imitation learning algorithms, including DT \cite{DT}, StARformer \cite{starformer}, and straightforward behavior cloning. 
We report results from the corresponding papers for CQL, REM, and QR-DQN.
For DT, there is a slight discrepancy between \cite{DT} and \cite{starformer}; we report raw data provided to us by DT authors.
As StARformer performs well on Atari, we use it as the main comparison object for GDT-plus.
In our behavior cloning setting, the agent lacks access to reward signals and online data from the environment, making the problem even more challenging. This differs from traditional imitation learning approaches that can collect new data and perform Inverse Reinforcement Learning \cite{abbeel2004apprenticeship, ng2000algorithms}. To create this setting, we remove the rewards from the dataset used in offline RL.

For the Gym environment, we compare GDT with a range of model-free methods, including CQL \cite{CQL}, BEAR \cite{BEAR}, BRAC \cite{BRAC}, and AWR \cite{AWR}, and model-based methods, including MBOP \cite{MBOP}. 
We also compare the performance of GDT with several imitation learning algorithms, including DT \cite{DT}, StARformer \cite{starformer}, and BC. 
Results for CQL, MBOP, and DT are reported from the original papers, while the results of BEAR, BRAC, and AWR are reported from the D4RL paper. 
Note that we re-run the experimental results on Gym for StARformer based on the official code published by the authors, as the original paper used the DeepMind Control Suite (DMC) environment with image input instead of vector input like Gym.

\subsection{Training Resources}

We use one NVIDIA A100 GPU (SXM4) to train each model.
Training each model typically takes 8-20 hours. 
However, since each environment needs to be trained three times with different seeds, the total training time is usually multiplied by three.

\section{Detailed results}
Table \ref{tab:baseline} presents the normalized scores used for normalization, as proposed in \cite{hafner2020mastering, d4rl}. 
Tables \ref{tab:raw_atari} and \ref{tab:raw_d4rl} show the raw scores corresponding to Tables \ref{tab:atari} and \ref{tab:d4rl}. 
We note a slight discrepancy between \cite{REM} and \cite{CQL} for REM and QR-DQN, and report the raw data provided to us by the REM authors. 
To alleviate the complexity of presenting a large amount of experimental data, we report only the mean in the tables. 

\begin{table}[h]
\caption{Baseline scores used for normalization.}
 \label{tab:baseline}
 \vskip 0.15in
\centering
\small
\begin{tabular}{lrr}
\toprule
\multicolumn{1}{c}{\bf Game} & \multicolumn{1}{c}{\bf Random} & \multicolumn{1}{c}{\bf Gamer} \\
  \midrule
Breakout & $2$ & $30$ \\
Qbert & $164$ & $13455$ \\ 
Pong & $-21$ & $15$ \\
Seaquest & $68$ & $42055$ \\
HalfCheet & $-280.2$ & $12135$ \\
Hopper & $-20.3$ & $3234.3$ \\
Walker & $1.6$ &	$4592.3$ \\
 \bottomrule
 \end{tabular}
\end{table}


\begin{table*}[h]
\centering
\caption{
Raw scores for 1\% DQN-replay Atari datasets. We evaluate the performance of GDT on four Atari games using three different seeds, and report the mean and variance of the results. 
Best mean scores are highlighted in bold.
The assessment reveals that GDT surpasses conventional RL algorithms on most tasks and achieves better performance than DT across all games. 
In contrast, GDT-plus achieves comparable results to StAR, leveraging fine-grained spatial information.}
\label{tab:raw_atari}
 \vskip 0.15in
\scalebox{0.85}{
\begin{tabular}{lrrrrrrrr}
\toprule
\multicolumn{1}{c}{\bf Game} & \multicolumn{1}{c}{\bf CQL} & \multicolumn{1}{c}{\bf QR-DQN} & \multicolumn{1}{c}{\bf REM} & \multicolumn{1}{c}{\bf BC}  & \multicolumn{1}{c}{\bf DT} & \multicolumn{1}{c}{\bf StAR} & \multicolumn{1}{c}{\bf GDT } & \multicolumn{1}{c}{\bf GDT-plus} \\
\midrule
Breakout  & $61.1$ & $6.8$ & $4.5$& $40.9 \pm 17.3$ & $76.9 \pm 27.3$ & $124.1 \pm 19.8$ &	$112.2 \pm 7.7$	& $\bf{125.7} \pm 3.2$ \\
Qbert     & $\bf{14012.0}$ & $156.0$ & $160.1$ & $2464.1 \pm 1948.2$ & $2215.8 \pm 1523.7$ & $6968.0 \pm 1698.0$ & $6205.8 \pm 2104.5$ & $7029.2 \pm 2928.5$ \\
Pong      & $\bf{19.3}$ & $-14.5$ & $-20.8$ & $9.7 \pm 7.2$ & $17.1 \pm 2.9$ & $18.9 \pm 0.7$&	$18.0 \pm 0.6$&	$19.0 \pm 0.1$ \\
Seaquest  & $779.4$ & $250.1$  & $370.5$   & $968.6 \pm 133.8$ & $1129.3 \pm 189.0$ & $781 \pm 212.0$	& $\bf{1261.3} \pm 204.3$ &	$1188 \pm 387.4$ \\
\bottomrule
\end{tabular}}
\end{table*}

\begin{table*}[tb!]
\centering
\caption{Raw scores for D4RL datasets. The performance of GDT is evaluated using three different seeds, and the mean is reported. 
Best mean scores are highlighted in bold.
The results demonstrate that GDT exhibits superior performance compared to conventional RL algorithms on most of the evaluated tasks, and GDT-plus performs comparably to GDT.}
\label{tab:raw_d4rl}
\vspace{.2cm}
\scalebox{0.75}{%
\begin{tabular}{>{\centering}p{0.14\textwidth}>{\centering}p{0.10\textwidth}>{\centering}p{0.06\textwidth}>{\centering}p{0.06\textwidth}>{\centering}p{0.08\textwidth}>{\centering}p{0.08\textwidth}>{\centering}p{0.06\textwidth}>{\centering}p{0.06\textwidth}>{\centering}p{0.10\textwidth}>{\centering}p{0.10\textwidth}>{\centering}p{0.10\textwidth}>{\centering\arraybackslash}p{0.10\textwidth}}
\toprule
Dataset       & Environment & CQL    & BEAR   & BRAC-v & AWR    & MBOP   & BC     & DT     & StAR   & GDT    & GDT-p  \\ \midrule
Medium-Expert & HalfCheetah & 7466.9 & 6349.5 & 4921.8 & 6262.6 & \bf{12867.5} & 7156.5 & 10496.2 & 11352.8 & 11191.4 & 11290.8 \\
Medium-Expert & Hopper      & 3592.3 & 3113.9 & 5.8    & 861.7  & 1773   & 2570.4 & 3481.6 & \bf{3595.6} & 3589   & \bf{3595.6} \\
Medium-Expert & Walker      & 4532.6 & 1842.5 & 3747.6 & 2471.4 & 3224.3  & 1681.8 & 4964.1  & 5005.5  & \bf{5019.2}  & 4945.8  \\ \midrule
Medium        & HalfCheetah & 5232.2 & 4897   & \bf{5468}   & 4363.1 & 5257   & 5070.8 & 5008.7 & 5045.9 & 5045.9 & 5045.9 \\
Medium        & Hopper      & 1867.4 & 1675.4 & 991.9  & 1148.1 & 1568   & 2059.4 & 2179.8 & 1916.2 & 2121.2 & \bf{2489}   \\
Medium        & Walker      & 3637.4 & 2714.7 & \bf{3724.7} & 800.4  & 1883.8 & 3550.2 & 3398.7 & 3389.5 & 3573.2 & 3513.5 \\ \midrule
Medium-Replay & HalfCheetah & 5455.6 & 4512.1 & \bf{5641.9} & 4723.1 & 4971.4  & 253.7  & 4263.8  & 4288.6  & 4673.5  & 4748    \\
Medium-Replay & Hopper      & 1561.4 & 1076.5 & -0.7   & 904    & 383.3  & 878    & 2671.3 & 930.1  & 2635.5 & \bf{2755.9} \\
Medium-Replay & Walker      & 1227.3 & 883    & 42.9   & 713.2  & 446.9  & 1695.6 & 3059   & 1828.7 & 3435.5 & \bf{3559.4}  \\ \bottomrule
\end{tabular}%
}
\end{table*}

\section{Hyper-parameters}

Tables \ref{tab:gdtatari}, \ref{tab:gdtpatari}, \ref{tab:gdtgym}, and \ref{tab:gdtpgym} provide a comprehensive list of hyper-parameters for our proposed GDT and GDT-plus models applied to Atari and OpenAI Gym environments. 
To ensure a fair comparison, we adopt similar hyper-parameter settings to Decision-Transformer \cite{DT}, including the number of Transformer layers, multi-head self-attention heads, and embedding dimensions in our Graph Transformer, as well as learning rate and optimizer configurations.

\begin{table*}[ht]
\caption{Hyperparameters of GDT for Atari experiments.}
\label{tab:gdtatari}
\vskip 0.15in
\begin{center}
\begin{small}
\begin{tabular}{ll}
\toprule
\textbf{Hyperparameter} & \textbf{Value}  \\
\midrule
Layers (Graph Transformer) & $6$  \\ 
MSA heads (Graph Transformer)   & $8$  \\
Embedding dimension (Graph Transformer)     & $128$  \\ 
Batch size   & $128$ \\
Context length $K$ & $50$ Pong \\
& $30$ Breakout, Qbert, Seaquest\\
Return-to-go conditioning & $120$ Breakout \\
& $5000$ Qbert  \\
& $20$ Pong  \\
& $1450$ Seaquest  \\
Nonlinearity & ReLU, encoder \\
& GeLU, otherwise \\
Max epochs & $10$ \\
Dropout & $0.1$ \\
Learning rate & $6*10^{-4}$ \\
Adam betas & $(0.9, 0.95)$ \\
Grad norm clip & $1.0$ \\
Weight decay & $0.1$ \\
Learning rate decay & Linear warmup and cosine decay (see code for details) \\
Warmup tokens & $512*20$ \\
Final tokens & $6*500000*K$ \\
\bottomrule
\end{tabular}
\end{small}
\end{center}
\end{table*} 

\begin{table*}[ht]
\caption{Hyperparameters of GDT-plus for Atari experiments.}
\label{tab:gdtpatari}
\vskip 0.15in
\begin{center}
\begin{small}
\begin{tabular}{ll}
\toprule
\textbf{Hyperparameter} & \textbf{Value}  \\
\midrule
Layers (Graph Transformer) & $6$  \\ 
MSA heads (Graph Transformer)   & $8$  \\
Embedding dimension (Graph Transformer)     & $128$  \\ 
Connection method & Stack \\
Layers (Sequence Transformer) & $2$  \\ 
Image patch size & 14 \\
MSA heads (Sequence Transformer)   & $4$  \\
Embedding dimension (Sequence Transformer)     & $64$  \\ 
Batch size   & $64$ \\
Context length $K$ & $50$ Pong \\
& $30$ Breakout, Qbert, Seaquest\\
Return-to-go conditioning & $120$ Breakout \\
& $5000$ Qbert  \\
& $20$ Pong  \\
& $1450$ Seaquest  \\
Nonlinearity & ReLU, encoder \\
& GeLU, otherwise \\
Max epochs & $10$ \\
Dropout & $0.1$ \\
Learning rate & $6*10^{-4}$ \\
Adam betas & $(0.9, 0.95)$ \\
Grad norm clip & $1.0$ \\
Weight decay & $0.1$ \\
Learning rate decay & Linear warmup and cosine decay (see code for details) \\
Warmup tokens & $512*20$ \\
Final tokens & $10*500000*K$ \\
\bottomrule
\end{tabular}
\end{small}
\end{center}
\end{table*} 

\begin{table*}[ht]
\caption{Hyperparameters of GDT for OpenAI Gym experiments.}
\label{tab:gdtgym}
\vskip 0.15in
\begin{center}
\begin{small}
\begin{tabular}{ll}
\toprule
\textbf{Hyperparameter} & \textbf{Value}  \\
\midrule
Layers (Graph Transformer) & $6$  \\ 
MSA heads (Graph Transformer)    & $8$  \\
Embedding dimension (Graph Transformer)    & $128$  \\
Nonlinearity function & ReLU \\
Batch size   & $64$ \\
Context length $K$ & $20$ \\
Return-to-go conditioning   & $6000$ HalfCheetah \\
                            & $3600$ Hopper \\
                            & $5000$ Walker \\
Dropout & $0.1$ \\
Learning rate & $10^{-4}$ \\
Grad norm clip & $0.25$ \\
Weight decay & $10^{-4}$ \\
Learning rate decay & Linear warmup for first $10^5$ training steps \\
\bottomrule
\end{tabular}
\end{small}
\end{center}
\end{table*} 

\begin{table*}[ht]
\caption{Hyperparameters of GDT-plus for OpenAI Gym experiments.}
\label{tab:gdtpgym}
\vskip 0.15in
\begin{center}
\begin{small}
\begin{tabular}{ll}
\toprule
\textbf{Hyperparameter} & \textbf{Value}  \\
\midrule
Layers (Graph Transformer) & $6$  \\ 
MSA heads (Graph Transformer)    & $8$  \\
Embedding dimension (Graph Transformer)    & $128$  \\
Connection method & Fusion \\ 
Layers (Sequence Transformer) & $6$  \\ 
MSA heads (Sequence Transformer)    & $8$  \\
Embedding dimension (Sequence Transformer)    & $256$  \\
Nonlinearity function & ReLU \\
Batch size   & $64$ \\
Context length $K$ & $20$ \\
Return-to-go conditioning   & $6000$ HalfCheetah \\
                            & $3600$ Hopper \\
                            & $5000$ Walker \\
Dropout & $0.1$ \\
Learning rate & $10^{-4}$ \\
Grad norm clip & $0.25$ \\
Weight decay & $10^{-4}$ \\
Learning rate decay & Linear warmup for first $10^5$ training steps \\
\bottomrule
\end{tabular}
\end{small}
\label{tbl:gym_hyperparameters}
\end{center}
\end{table*} 

\end{document}